\documentclass[letterpaper, 10 pt, conference]{ieeeconf} 
\IEEEoverridecommandlockouts                              
\usepackage{url}
\usepackage{amsmath,amsfonts}
\usepackage{array}
\usepackage{textcomp}
\usepackage{stfloats}
\usepackage{verbatim}
\usepackage{graphicx}
\usepackage{cite}
\usepackage{color}
\usepackage{colortbl}
\usepackage{soul}
\usepackage{multirow}
\usepackage{multicol}
\usepackage{makecell}
\usepackage{booktabs}
\usepackage{hyperref}
\usepackage{booktabs}
\usepackage{amssymb}
\usepackage{bbding}
\usepackage{pifont}
\usepackage{wasysym}
\usepackage{subfigure}
\usepackage{amsfonts}
\usepackage{pifont}
\usepackage{amsmath}
\usepackage{bm}
\usepackage[table]{xcolor}
\usepackage[ruled,vlined]{algorithm2e}
\usepackage{algpseudocode}
\usepackage[utf8]{inputenc}
\usepackage{CJK}
\usepackage{amsmath}
\usepackage{threeparttable}
\newcolumntype{P}[1]{>{\centering\arraybackslash}p{#1}}
\newcolumntype{C}[1]{>{\centering\arraybackslash}m{#1}}
\newcommand{\rt}{\textcolor[rgb]{1,0,0}}
\newcommand{\bt}{\textcolor[rgb]{0,0,1}}

\newcommand{\gt}{\textcolor[rgb]{0,1,0}}

\begin{document}

\title{\LARGE \bf
    LiteVLoc: Map-Lite Visual Localization for Image Goal Navigation
}

\author{
    Jianhao Jiao$^{1}$, Jinhao He$^{2}$, Changkun Liu$^{3}$, Sebastian Aegidius$^{1}$, Xiangcheng Hu$^{4}$,\\
    Tristan Braud$^{3}$, Dimitrios Kanoulas$^{1, 5}$
    \thanks{\textsuperscript{1}The authors are with the Robot Perception and Learning Lab, Intelligent Robotics, Department of Computer Science, University College London, Gower Street, WC1E 6BT, London, UK. {\tt\small \{ucacjji, d.kanoulas\}@ucl.ac.uk}.}
    \thanks{\textsuperscript{2}The author is with the Thrust of Robotics and Autonomous Systems, The Hong Kong University of Science and Technology (Guangzhou), Nansha District, Guangzhou, China.
    }
    \thanks{\textsuperscript{3}The authors are with the Department of Computer Science and Engineering, HKUST, Hong Kong, China.
    }
    \thanks{\textsuperscript{4}The author is with the Department of Electronic an Computer Engineering, HKUST, Hong Kong, China.
    }
    \thanks{\textsuperscript{5}Dimitrios Kanoulas is also with the AI Centre, Department of Computer Science, University College London, Gower Street, WC1E 6BT, London, UK and Archimedes/Athena RC, Greece.}
    \thanks{This work was supported by the UKRI Future Leaders Fellowship [MR/V025333/1] (RoboHike).  For the purpose of Open Access, the author has applied a CC BY public copyright license to any Author Accepted Manuscript version arising from this submission.}}

\maketitle

\begin{abstract}
This paper presents LiteVLoc, a hierarchical visual localization framework that uses a lightweight topo-metric map to represent the environment. The method consists of three sequential modules that estimate camera poses in a coarse-to-fine manner. Unlike mainstream approaches relying on detailed 3D representations, LiteVLoc reduces storage overhead by leveraging learning-based feature matching and geometric solvers for metric pose estimation. A novel dataset for the map-free relocalization task is also introduced. Extensive experiments including localization and navigation in both simulated and real-world scenarios have validate the system's performance and demonstrated its precision and efficiency for large-scale deployment. Code and data will be made publicly available at \url{https://rpl-cs-ucl.github.io/LiteVLoc}.
\end{abstract}
\section{Introduction}
\label{sec:introduction}


\subsection{Motivation}
\label{sec:motivation}

Imagine that you start the daily route to campus from the nearest subway station.
Before setting out, it is essential to determine your relative position and orientation to the destination.
As you commence, by iteratively repeating this process, you should refine and update your location as well as adjust the planned route.
This strategy will ensure the efficient and reliable navigation to the destination.

The above process of determining the position and orientation of a body relative to a reference frame using images is known as visual localization (VLoc).
Most of VLoc methods \cite{noh2017large,dusmanu2019d2,sattler2016efficient,taira2018inloc,sarlin2019coarse} follow a standard procedure that initially constructs a metric map of the environment and then predict the camera's poses within this map.
To create a metrically precise map, VLoc methods typically employ off-the-shelf Structure-from-Motion (SfM) software \cite{schoenberger2016sfm}, which ensures accurate mapping.
Alternatively, less refined maps can be constructed in real time using SLAM, though this approach often comes at the cost of reduced accuracy.
In either case, the resulting maps consume vast amounts of storage \cite{brachmann2023accelerated} (e.g., $2$\textit{GB} for an indoor scene of hLoc \cite{sarlin2019coarse}), and present challenges for maintenance, particularly in the context of long-term navigation.
As Heraclitus famously stated, `You cannot step into the same river twice'; the concept is further underscored in a recent survey \cite{yin2024general}, highlighting that environments are subject to continual change in reality.
This introduces a paradox: the more detailed and precise the description of an environment, the less adaptable it becomes when the environment inevitably shifts.


The aim of this paper is to develop a VLoc method that delivers camera poses at a metric level using a lightweight and sparse map.
Previous research has demonstrated that the topo-metric map is a favorable representation due to its simplicity, flexibility, and scalability \cite{dayoub2013vision, roussel2020deep, rosinol2021kimera, he2023metric, kim2023topological, hughes2024foundations}.
This map formulates the environment as a graph with nodes that contain images or visual features and spatial coordinates, and edges that represent node relationships.
Despite these benefits, the adoption of topo-metric maps in real-world robotic systems remains limited, likely due to their sparse nature, which poses challenges for VLoc.

\subsection{Contributions}
\label{sec:contribution}
In this paper, we propose \textbf{LiteVLoc}, a hierarchical and map-lite VLoc method for robot navigation by exploiting the topo-metric map.
LiteVLoc, which is inspired from the hierarchical framework of VLoc \cite{sarlin2019coarse},
estimates the camera's poses in a coarse-to-fine manner with three successive modules:
\textit{(1)} the Global Localization (GL) module initializes the camera's poses at a topological level,
\textit{(2)} the Local Localization (LL) module refines the camera's pose at a metric level, and
\textit{(3)} the Pose SLAM (PS) module integrates the low-rate VLoc's results with high-rate sensor-specific odometry to deliver real-time poses in the global coordinate for subsequent navigation tasks.

The LL module attempts to solve the challenge of map-free relocalization \cite{arnold2022map}, which involves estimating a query pose from a single reference image.
This task becomes particularly difficult when significant differences in appearance, viewpoint, or relative distance occur between the query and the reference images.
We introduce a new dataset specifically for this task, along with a detailed evaluation of state-of-the-art (SoTA) feature matching models.
Our findings reveal that the highest-scoring method can establish reliable correspondences between images across diverse environments on a zero-shot basis, significantly enhancing our method's generalization capabilities.

In addition to evaluating LiteVLoc on VLoc tasks, we have deployed it on a real-world legged robot to perform image-goal visual navigation (VNav) tasks.
This integration enables better human-robot interaction by providing a more intuitive navigation goal to the robot, unlike the traditional method of specifying goals with coordinates.
In general, this paper presents the following \textit{contributions}:
\begin{enumerate}
  \item We propose a general, scalable, and hierarchical VLoc framework utilizing a topo-metric map for robot navigation. This framework's \textit{generalization} facilitates adaptation to various robots with specific vision-based approaches, while its \textit{scalability} reduces reliance on specialized mapping devices, highlighting its potential for large-scale deployment.
  \item Within this framework, we present a specific VLoc on a real-world robot.
        The map can be constructed from various geo-tagged image sources, including robot-mounted cameras and mobile devices like AR glasses \cite{engel2023project}, despite existing domain gaps.
        The map size can be reduced to $17$\textit{MB} covering a $200m$ route, with an average translation error less than $0.25m$.
  \item We introduce a new map-free relocalization dataset, collected by mobile robots in environments ranging from simulated offices to real-world campuses. As a supplement to the existing dataset \cite{arnold2022map}, ours offers greater diversity in viewpoints and relative distances. This resource is vital for evaluating map-free relocalizers for mobile robot applications.
  \item We further implement and verify the image-goal VNav system in both simulated and real-world environments based on the map and VLoc method.
\end{enumerate}
\section{Related Work}
\label{sec:related_work}
Environmental representations for VLoc can be broadly classified into categories based on whether the approaches employ a pre-built 3D metric map (i.e., with or without 3D representation) to estimate the camera's poses.



\subsection{VLoc without 3D Representation}
The simplest form of a map consists of a database of visual features or mapping images and their corresponding poses. When the image nodes are connected, this map can be structured as a topological or topo-metric map. Given a query image, the most similar images from the map are retrieved~\cite{arandjelovi2015netvlad,xu2021probabilisticvp,berton2022rethinking,berton2023eigenplaces,gordo2017end}, and the query pose is estimated based on the poses of the top retrieved images~\cite{sattler2019understanding,dong2023lazy,zaffar2024estimation}. These methods, commonly referred to as image retrieval or visual place recognition (VPR), discretize the environment into distinct places but cannot provide fine-grained estimates of the 6-DOF pose, often resulting in large error margins (e.g., $5$\,m and $90^\circ$ as reported in~\cite{xu2021probabilisticvp}).
An alternative approach, pose regression~\cite{kendall2015posenet,kendall2017geometric,shavit2021learning,chen2022dfnet,balntas2018relocnet,laskar2017camera,sattler2019understanding}, trains scene-specific neural networks to predict the absolute or relative pose directly from a query image using the database. While pose regression is computationally efficient and fast, it generally lacks the accuracy and generalization ability of structure-based methods~\cite{sattler2019understanding,10610903}.

\subsection{VLoc with 3D Representation}
Structure-based approaches traditionally rely on sparse 3D models that are generated from SfM to establish 2D-3D correspondences~\cite{detone2018superpoint,lindenberger2023lightglue,sarlin2020superglue,sun2021loftr,sarlin2019coarse}.  After obtaining 2D-3D correspondences,  the camera pose can be estimated via Perspective-n-Point (PnP)~\cite{gao2003complete}. Alternatives to SfM models include meshes~\cite{panek2022meshloc}, Neural Radiance Fields (NeRFs)~\cite{moreau2023crossfirecr,liu2023nerfloc,zhou2024nerfect}, and 3D Gaussian Splatting~\cite{liu2024gsloc}. However, generating high-quality 3D representation in complex environments is often difficult and time-consuming~\cite{brachmann2023accelerated}.
Scene coordinate regression (SCR) methods directly predict 3D coordinates for given 2D pixel positions in the query image~\cite{brachmann2023accelerated,brachmann2016dsac,brachmann2021dsacstar,brachmann2018lessmore}, which can alleviate this issue.
Despite this, SCR requires training a scene-specific model and are typically insufficiently robust or accurate for large-scale scenes~\cite{brachmann2023accelerated}.

LiteVLoc falls within the VLoc without 3D representation category, utilizing a topo-metric map to represent the environment.
This eliminates the need to construct a 3D metrically-precise map for localization and streamlines the mapping process for large-scale navigation tasks.
Different from VPR, LiteVLoc employs robust feature matching to establish correspondences between the query image and the top-retrieved image.
By using depth information from stereo cameras, our approach can achieve localization results with accuracy comparable to that of using dense 3D representations.s
In contrast to pose regression and SCR methods, LiteVLoc is scene-agnostic and can be applied to unseen environments without requiring network fine-tuning.

\section{Methodology}
\label{sec:methodology}
This section presents the detailed methodology of the LiteVLoc framework and the VNav system derived from it.

\begin{figure}[t]
  \begin{center}
    \includegraphics[width=0.91\linewidth]{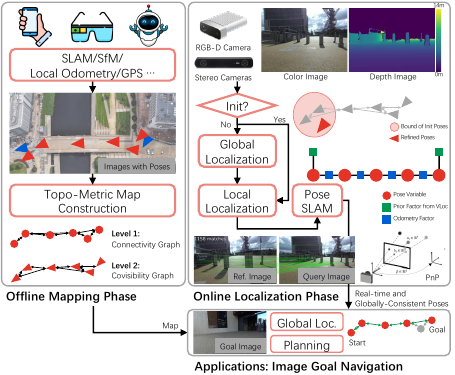}
  \end{center}
  \vspace{-0.35cm}
  \caption{Pipeline of LiteVLoc.
    The mapping phase builds a topo-metric map from a set of images with poses (Section \ref{sec:mapping}).
    Selected cameras are marked in \rt{red}; otherwise, they are marked in \bt{blue}.
    The map includes two levels, with edges marked in black: level-$1$ for planning and level-$2$ for VLoc.
    The localization phase estimates camera's poses in a coarse-to-fine manner (Section \ref{sec:visual_localization}). The map and real-time localization results can be applied to various applications such as image goal navigation (Section \ref{sec:navigation}). The image to illustrate PnP is derived from \cite{sheffer2020pnp}.}
  \label{fig:overview_visual_loc}
  \vspace{-0.5cm}
\end{figure}

\subsection{Preliminary}

\subsubsection{Problem Definition}
Let $\mathbf{x} = [\mathbf{t}, \mathbf{q}]$ represent the camera's pose, where $\mathbf{t} = [x, y, z]^{\top}$ is the position, and $\mathbf{q} = [qw, qx, qy, qz]^{\top}$ is the quaternion representing rotation.
Without loss of generality, we consider that the robot frame coincides with the camera frame, using ``camera pose'' to refer to the ``robot pose.''
The camera's pose in the world frame at time step $k$ is denoted as $\mathbf{x}_{k}$.
Given the pre-built and unchanged map $\mathcal{M}$, visual observations $\mathcal{Z}_{k}$, interoceptive sensor measurements $\mathcal{U}_{k}$ (e.g., IMU data), and the prior pose $\hat{\mathbf{x}}_{k-1}$ from the last time step, we formulate the localization as an optimization problem \cite{barfoot2024state}
\begin{equation}
  \underset{\mathbf{x}_{k}}{\arg\min}\
  \underbrace{[
        f_{u}(\mathbf{x}_{k}, \mathcal{U}_{k}, \hat{\mathbf{x}}_{k-1})
        +
        f_{z}(\mathbf{x}_{k}, \mathcal{M}, \mathcal{Z}_{k})
      ]}
  _{f(\mathbf{x}_{k},\mathcal{M}, \mathcal{Z}_{k}, \mathcal{U}_{k}, \hat{\mathbf{x}}_{k-1})}
  .
  \label{equ:formulation_vloc_opt}
\end{equation}
where $f(\cdot)$ is decomposed into two objectives:
$f_u(\cdot)$, which measures the deviation between the propagated pose and the variable, and
$f_z(\cdot)$, which quantifies the deviation between the predicted measurements and practical observations.
But solving problem \eqref{equ:formulation_vloc_opt} is challenging due to its non-linear and non-convex nature.
To make it tractable, this paper approximates the solution by first addressing a vision-based localization and then tackling a pose SLAM problem.


\subsubsection{LiteVLoc Framework}
Figure \ref{fig:overview_visual_loc} illustrates the pipeline of the LiteVLoc's framework.
The process begins with the GL module, which initializes the camera's coarse position in the global coordinate system defined by the map.
It is triggered when the mission starts or the robot loses its position (e.g., sufficient feature correspondences cannot be established over an extended period).
Once initialized, the LL module refines the camera's pose at a metric level.
It consist of three steps: searching for the most similar reference node, establishing correspondences between the reference image and the observation, and estimating the camera's pose.
Finally, the PS module fuses low-rate VLoc's results with local odometry (e.g.,  proprioceptive odometry), which provides high-rate and accurate pose estimates.

\subsection{Map Construction}
\label{sec:mapping}
\subsubsection{Map Representation}
We design a two-level topo-metric map to support VLoc and VNav.
Each level is represented as an undirected graph $\mathcal{G} = \{\mathcal{N}, \mathcal{E}\}$,
where $\mathcal{N}$ is the set of nodes and $\mathcal{E}$ is the set of edges depicting nodes' relationships.
The first level, the \textit{connectivity graph} (CnG), is used for motion planning. Nodes are connected if they are navigable from each other, and each node stores the pose.
The second level, the \textit{covisibility graph} (CvG), is used for VLoc. Nodes are connected if they visual overlaps with each other, and in addition to the pose, each node stores visual VPR features and raw images captured in this place.

\subsubsection{Keyframe Selection}
For simplicity, we assume the CnG is equal to the CvG.
We use the CvG as an example to explain how the map is constructed.
We assume that a set of visual observations and their associated poses which are denoted by \(\mathcal{C} = \{(\mathcal{Z}_i, \mathbf{x}_i)\}\) are available.
$\mathcal{C}$ can be obtained from various sources, such as single-/multi-agent SLAM or SFM algorithms \cite{xu2024d, liu2024marvin}, cameras mounted on robots with odometry (e.g., wheel encoders, GPS) \cite{shah2023gnm}, or internet databases (e.g., Google Street View) \cite{anguelov2010google}.
Covisibility between frames is determined by checking for sufficient feature correspondences.
To reduce the storage requirement, we select a subset of keyframes $\mathcal{\mathcal{C}^{\#}}$, while maximumly preserving the environmental information.
These keyframes with edges construct the CvG.
We formulate the keyframe selection as an optimization problem:
\begin{equation}
  \underset{\mathcal{\mathcal{C}^{\#}}}{\arg\max}\
  h(\mathcal{\mathcal{C}^{\#}}),
  \ \ \
  \text{subject to}
  \ |\mathcal{\mathcal{C}^{\#}}|\leq M,
  \ \mathcal{\mathcal{C}^{\#}}\subseteq\mathcal{C},
\end{equation}
where $h(\cdot)$ quantifies the information and \(M\) is a hyper-parameter.
We apply the greedy-based method \cite{jiao2021greedy} to approximate $\mathcal{C}^{\#}$ within a reasonable computation time.
If visual observations include depth images, we represent the environment using a 2D occupancy map constructed from selected keyframes, with $h(\cdot)$ defined as the number of occupied grid cells.
If depth images are not available, we use a simple approach by treating frames as point clouds and downsampling them at a fixed resolution to select keyframes.

\subsection{Coarse-to-Fine Visual Localization}
\label{sec:visual_localization}
LiteVLoc is algorithm-agnostic, meaning that it can be applied to virtually any VPR and feature matching model. In this section, we present a specific solution.

\subsubsection{Global Localization}
A key assumption in VPR is that geographically close places share similar layouts, which can be partially observed by cameras \cite{yin2024general}.
This \textit{similarity} can be reflected in the visual resemblance between two images and can be quantified by the relative distance between their \textit{global descriptors} (describing overall features of an image).
Following this assumption, we can initialize the camera's pose using that of the top-$1$ retrieved image if the current observation is similar to it.
We use the pre-trained CosPlace \cite{berton2022rethinking} model for VPR due to its high performance and simple architecture.
CosPlace consists of a convolutional encoder and a generalized mean pooling (GeM) layer to compute the global descriptor.
We use CosPlace with this configuration in experiments:
ResNet-18 \cite{he2016deep} as the backbone; the dimension of the descriptor is $256$.

\subsubsection{Local Localization}
Using the pose from the previous time step as a prior, we can retrieve nearby candidate nodes by first finding the closest map node.
Then, using covisibility cues from the CvG, we gather the map and its neighbors to form a set of candidates.
The most similar node from the set is selected as the reference node.

The following steps is to first establish local feature correspondences between the reference image (denoted by $\mathbf{I}_{ref}$) and the current observation, and then estimate the camera's pose.
The observation, denoted by $\mathcal{Z}_{k} = \{\mathbf{I}_{k}, \mathbf{D}_{k}\}$, offers a color and depth image.
Dense $2$D-$2$D pixel correspondences between $\mathbf{I}_{ref}$ and $\mathbf{I}_{k}$ are obtained using the MASt3R model \cite{leroy2024grounding}.
MASt3R, which was trained on extensive data, directly predicts $3$D geometry from image pairs and has demonstrated strong robustness in our tests.
We then obtain $2$D-$3$D correspondences by lifting $2$D pixels to $3$D points using the depth image $\mathbf{D}_{k}$ and the camera intrinsics.
Finally, the camera's pose is estimated by feeding these correspondences into a PnP solver \cite{gao2003complete} with a RANSAC loop \cite{fischler1981random}.
The number of inliers can be used to check whether the LL is successful.

\subsubsection{Pose SLAM}
VLoc's performance may be affected by several factors such as
\textit{(1)} large domain gaps and viewpoint differences between the map and observations, 
\textit{(2)} textureless or repetitive scenes, 
\textit{(3)} noise in depth data, and 
\textit{(4)} time delays in visual processing. 
While using inlier counts to filter out unreliable results can mitigate these issues, it reduces the update rate. 
To address this, we designed a PS module that integrates outputs from VLoc and local odometry, which enhances the system's overall reliability and performance.

Local odometry can come from possible sources that provide cumulative motion estimates, such as wheel encoders or visual odometry.
In simulations, we implement an ICP algorithm for depth point cloud \cite{rusinkiewicz2001efficient}, while in real-world environments, we use proprioceptive odometry \cite{bloesch2017state} that mixes IMU, encoder and robot contact data.
We model the PS problem as a factor graph (Fig. \ref{fig:overview_visual_loc}) and solve it using the GTSAM library \cite{framk2022gtsam} from the optimization perspective.
With camera's poses as variables, the cost function includes error terms from both prior and odometry constraints. 
The prior constraints, provided by VLoc, offer initial estimates of camera's poses relative to the map, while the odometry constraints, derived from local odometry, provide estimates of pose changes at each time step.
Optimization is triggered with new VLoc results; otherwise, the pose is directly propagated by the local odometry.

\subsection{Closed-Loop Visual Navigation}
\label{sec:navigation}
We integrate LiteVLoc with motion planning to achieve the closed-loop image goal VNav (see Fig. \ref{fig:overview_visual_loc}).
Unlike point-goal navigation, where the target is specified by coordinates, image-goal navigation directs the robot to a target image \cite{shah2023gnm}, enhancing better human-robot interaction.
Although this paper shows the image-goal visual navigation, the proposed approach can be extended to tasks such as object-goal or instruction-based navigation.
By leveraging the map and VLoc method, we can localize the goal using VPR.

Planning is hierarchical, with global planning using Dijkstra algorithm to find the shortest path from the start to the goal node on the CnG. Traversed nodes are stored as subgoals.
By projecting the nearest subgoal's coordinates onto the robot frame, local motion planning generates a collision-free path to this goal using single-frame depth measurements and the robot's kinematics,
In simulations, we use iPlanner \cite{yang2023iplanner}.
To avoid fine-tuning in real-world tests, we modify the SoTA nonlearning motion primitives planner (Falco) \cite{zhang2020falco}, which requires only a single-frame depth point cloud as input and eliminates the need to build a local map.


\section{Experiment}
\label{sec:experiment}

\begin{figure}[t]
  \centering
  \subfigure[Sample images of different scenarios: (top) Matterport3D, (middle) UCL Campus, and (bottom two rows) HKUSTGZ Campus.]
  {\label{fig:xxx}\centering\includegraphics[width=0.99\linewidth]{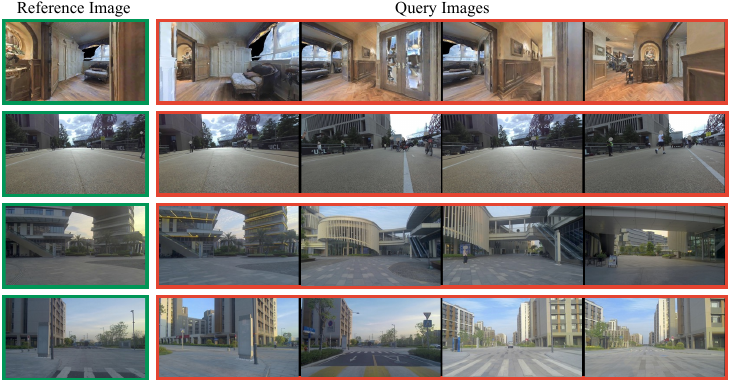}}
  \subfigure[Poses of cameras (arrows indicate the camera's front view). Green and red arrows represent reference images and query images, respectively.]
  {\label{fig:xxxxx}\centering\includegraphics[width=0.99\linewidth]{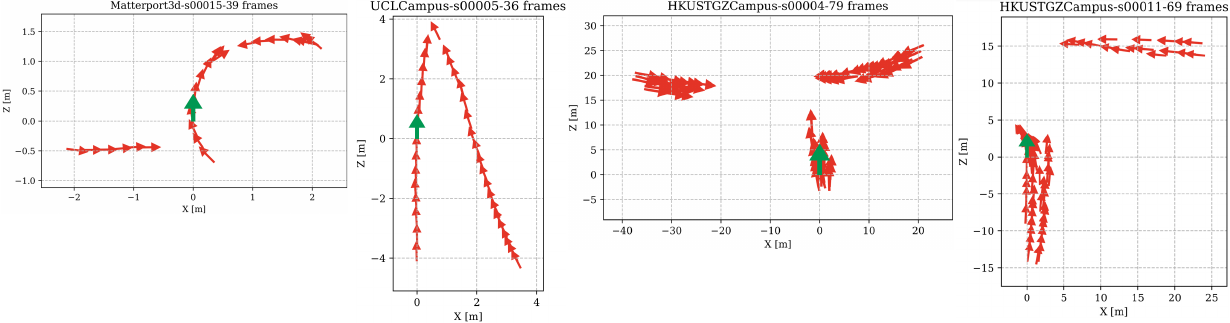}}
  \caption{Sample images and the distribution of camera poses in the proposed dataset for evaluating map-free relocalization methods.}
  \label{fig:exp_rpe_benchmark}
  \vspace{-0.5cm}
\end{figure}

\begin{table*}[]
  \centering
  \caption{Performance of relative pose estimation with different feature matchers.
    $[Xcm, Y^{\circ}]$ indicates the precision of pose error within this range. The first and second best results are highlighted in \rt{red} and \bt{blue}.}
  \renewcommand\arraystretch{0.95}
  \renewcommand\tabcolsep{1.0pt}
  \scriptsize
  \vspace{-0.17cm}
  \begin{tabular}{c|c|c|c|c|c|c|c|c}
    \multicolumn{9}{c}{\multirow{1}{*}{\textbf{\footnotesize{Simulated Matterport3D $+$ UCL Campus Dataset}}}}                                                                                                                                                                                                                                                                                                   \\
    \toprule[0.03cm]
    \textbf{Types} & \textbf{Methods}                                                           & \textbf{Max} $E_{t}[m]/E_{r}[deg]\downarrow$ & \textbf{Avg. Median} $E_{t}[m]/E_{r}[deg]\downarrow$ & $[\mathbf{5}cm, \mathbf{5}^\circ]\uparrow$ & $[\mathbf{25}cm, \mathbf{5}^\circ]\uparrow$ & $[\mathbf{1}m, \mathbf{10}^\circ]\uparrow$ & \textbf{\% Estimated} $\uparrow$ & \textbf{Time}$[ms]\downarrow$ \\
    \midrule[0.03cm]
    \multirow{4}{*}{\textbf{Dense}}
                   & RoMa \cite{edstedt2024roma}                                                & $6.52/169.75$                                & $0.26/0.73$                                          & $62.44$                                    & $75.58$                                     & $86.85$                                    & $\rt{\mathbf{100}}$              & $321$                         \\
                   & DUSt3R \cite{wang2024dust3r}                                               & $6.55/\bt{\mathbf{43.85}}$                   & $\bt{\mathbf{0.16}}/1.25$                            & $40.84$                                    & $\bt{\mathbf{77.46}}$                       & $\bt{\mathbf{94.83}}$                      & $99.53$                          & $669$                         \\
                   & MASt3R \cite{leroy2024grounding}                                           & $5.78/\rt{\mathbf{14.85}}$                   & $\rt{\mathbf{0.06}}/\rt{\mathbf{0.64}}$              & $\bt{\mathbf{66.19}}$                      & $\rt{\mathbf{80.75}}$                       & $\rt{\mathbf{94.83}}$                      & $97.18$                          & $101$                         \\ \midrule
    \multirow{4}{*}{\textbf{\begin{tabular}[c]{@{}c@{}}Semi-\\ Dense\end{tabular}}}
                   & LoFTR \cite{sun2021loftr}                                                  & $\rt{\mathbf{4.65}}/179.31$                  & $0.42/4.38$                                          & $53.05$                                    & $69.95$                                     & $81.22$                                    & $\rt{\mathbf{100}}$              & $20$                          \\
                   & Efficient-LoFTR \cite{wang2024efficient}                                   & $7.01/178.94$                                & $0.45/8.56$                                          & $56.80$                                    & $69.01$                                     & $80.75$                                    & $\rt{\mathbf{100}}$              & $16$                          \\
                   & MatchFormer \cite{wang2022matchformer}                                     & $\bt{\mathbf{4.73}}/178.70$                  & $0.48/7.04$                                          & $46.00$                                    & $67.60$                                     & $77.93$                                    & $\rt{\mathbf{100}}$              & $36$                          \\
                   & XFeat*\cite{potje2024xfeat}                                                & $8.31/178.71$                                & $0.57/10.10$                                         & $28.16$                                    & $63.38$                                     & $76.52$                                    & $\rt{\mathbf{100}}$              & $\bt{\mathbf{12}}$            \\ \midrule
    \multirow{6}{*}{\textbf{Sparse}}
                   & SuperPoint-LightGlue \cite{detone2018superpoint,lindenberger2023lightglue} & $8.61/178.93$                                & $0.44/5.95$                                          & $49.29 $                                   & $69.01 $                                    & $80.75 $                                   & $98.59 $                         & $21$                          \\
                   & GIM-LightGlue \cite{shen2024gim,lindenberger2023lightglue}                 & $9.17/177.81$                                & $0.29/1.41$                                          & $36.15$                                    & $63.38$                                     & $78.40$                                    & $92.95$                          & $28$                          \\
                   & GIM-DKM \cite{shen2024gim,edstedt2023dkm}                                  & $7.94/175.43$                                & $0.20/\bt{\mathbf{0.69}}$                            & $65.72 $                                   & $79.81 $                                    & $90.61 $                                   & $\rt{\mathbf{100} }$             & $760 $                        \\
                   & XFeat \cite{potje2024xfeat}                                                & $5.66/176.88$                                & $0.57/11.58$                                         & $32.39$                                    & $60.09$                                     & $73.23$                                    & $\rt{\mathbf{100}}$              & $16$                          \\
                   & (Non-Learning) SIFT-NN \cite{lowe2004distinctive}                          & $427.31/171.37$                              & $2.93/6.49$                                          & $42.25$                                    & $63.38$                                     & $75.11$                                    & $90.14$                          & $24   $                       \\
                   & (Non-Learning) ORB-NN \cite{rublee2011orb}                                 & $207.71/179.52$                              & $0.49/8.11$                                          & $30.04$                                    & $55.86$                                     & $70.89$                                    & $90.14$                          & $\rt{\mathbf{11}}$            \\
    \bottomrule[0.03cm]
  \end{tabular}

  \vspace{0.13cm}

  \begin{tabular}{c|c|c|c|c|c|c|c|c}
    \multicolumn{9}{c}{\multirow{1}{*}{\textbf{\footnotesize{HKUSTGZ Campus Dataset}}}}                                                                                                                                                                                                                                                                                                                          \\
    \toprule[0.03cm]
    \textbf{Types} & \textbf{Methods}                                                           & \textbf{Max} $E_{t}[m]/E_{r}[deg]\downarrow$ & \textbf{Avg. Median} $E_{t}[m]/E_{r}[deg]\downarrow$ & $[\mathbf{5}cm, \mathbf{5}^\circ]\uparrow$ & $[\mathbf{25}cm, \mathbf{5}^\circ]\uparrow$ & $[\mathbf{1}m, \mathbf{10}^\circ]\uparrow$ & \textbf{\% Estimated} $\uparrow$ & \textbf{Time}$[ms]\downarrow$ \\
    \midrule[0.03cm]
    \multirow{4}{*}{\textbf{Dense}}
                   & RoMa \cite{edstedt2024roma}                                                & $114.49/179.90$                              & $2.69/8.80$                                          & $5.13$                                     & $46.15$                                     & $\bt{\mathbf{70.00}}$                      & $\rt{\mathbf{100}}$              & $312$                         \\
                   & DUSt3R \cite{wang2024dust3r}                                               & $\bt{\mathbf{58.38}}/\bt{\mathbf{175.23}}$   & $\bt{\mathbf{1.27}}/\bt{\mathbf{1.44}}$              & $3.85$                                     & $37.95$                                     & $67.44$                                    & $98.72$                          & $361$                         \\
                   & MASt3R \cite{leroy2024grounding}                                           & $\rt{\mathbf{52.65}}/\rt{\mathbf{174.91}}$   & $\rt{\mathbf{0.39}}/\rt{\mathbf{1.05}}$              & $\rt{\mathbf{5.39}}$                       & $\rt{\mathbf{47.44}}$                       & $\rt{\mathbf{74.36}}$                      & $96.92$                          & $99$                          \\
    \midrule
    \multirow{4}{*}{\textbf{\begin{tabular}[c]{@{}c@{}}Semi-\\ Dense\end{tabular}}}
                   & LoFTR  \cite{sun2021loftr}                                                 & $178.50/180$                                 & $3.20/10.86$                                         & $5.13$                                     & $38.97$                                     & $64.87$                                    & $\rt{\mathbf{100}}$              & $22$                          \\
                   & Efficient-LoFTR \cite{wang2024efficient}                                   & $86.32/179.58$                               &
    $2.96/11.08$   & $4.36$                                                                     & $41.28$                                      & $65.13$                                              & $99.49$                                    & $18$                                                                                                                                                        \\
                   & MatchFormer  \cite{wang2022matchformer}                                    & $189.29/179.84$                              & $3.03/8.65$                                          & $3.33$                                     & $37.95$                                     & $65.13$                                    & $98.97$                          & $37$                          \\
                   & XFeat*  \cite{potje2024xfeat}                                              & $262.07/179.90$                              & $3.17/10.35$                                         & $1.80$                                     & $22.56$                                     & $55.90$                                    & $94.62$                          & $\bt{\mathbf{14}}$            \\
    \midrule
    \multirow{6}{*}{\textbf{Sparse}}
                   & SuperPoint-LightGlue \cite{detone2018superpoint,lindenberger2023lightglue} & $204.06/179.94$                              & $1.82/7.28$                                          & $3.59$                                     & $39.49$                                     & $63.33$                                    & $91.54$                          & $23$                          \\
                   & GIM-LightGlue \cite{shen2024gim,lindenberger2023lightglue}                 & $148.17/179.64$                              & $2.24/7.04$                                          & $2.56$                                     & $35.64$                                     & $62.05$                                    & $89.49$                          & $33$                          \\
                   & GIM-DKM \cite{shen2024gim,edstedt2023dkm}                                  & $63.23/179.63$                               & $2.36/6.40$                                          & $\bt{\mathbf{5.64}}$                       & $\bt{\mathbf{47.18}}$                       & $\bt{\mathbf{70.00}}$                      & $\rt{\mathbf{100}}  $            & $800$                         \\
                   & XFeat \cite{potje2024xfeat}                                                & $84.16/179.84$                               & $4.36/17.30$                                         & $3.08$                                     & $27.69$                                     & $56.41$                                    & $\rt{\mathbf{100}}  $            & $19$                          \\
                   & (Non-Learning) SIFT-NN \cite{lowe2004distinctive}                          & $594.87/178.03$                              & $2.96/6.12$                                          & $2.05$                                     & $22.05$                                     & $45.39$                                    & $64.62$                          & $23$                          \\
                   & (Non-Learning) ORB-NN \cite{rublee2011orb}                                 & $607.47/179.84$                              & $1.55/1.84$                                          & $1.54$                                     & $16.15$                                     & $42.56$                                    & $69.49$                          & $\rt{\mathbf{13}}$            \\
    \bottomrule[0.03cm]
  \end{tabular}
  \label{tab:pose_error_evaluation}
  \vspace{-0.3cm}
\end{table*}

\subsection{Overview}
\label{sec:experiment_overview}

\subsubsection{Settings}
We conduct experiments from three perspectives, with increasing task complexity.
All experiments are evaluated using both simulated and real-world data.
\begin{itemize}
  \item \textbf{Map-free Relocalization} (Section \ref{sec:experiment_rpe}): We benchmark $13$ SoTA feature matchers coupled with a PnP $+$ RANSAC solver using a new dataset to evaluate pose estimation accuracy against severe viewpoint changes, following the setting in \cite{arnold2022map}.
  \item \textbf{Visual Localization} (Section \ref{sec:experiment_vloc}): We evaluate the complete LiteVLoc system with sequences captured from both simulated and real-world environments.
  \item \textbf{Image Goal Navigation} (Section \ref{sec:experiment_vnav}): We test the VNav system on both simulated and real-world robots, providing both quantitative and qualitative results.
\end{itemize}

\subsubsection{Implementation Details}
LiteVLoc is implemented in Python.
All algorithms, except for real-world VNav experiments, are run on an i$9$ desktop PC with an Nvidia GeForce RTX $4090$ GPU.
Real-world VNav experiments are conducted on ANYmal-D \cite{Hutter2016ANYmalA} quadruped robots, equipped with a front-view ZED$2$ stereo camera ($540\times 960$ resolution, generating reliable depth up to $15m$),
an Intel NUC computer (for planning), and a NVIDIA Jetson Orin (for visual processing).
Ground-truth (GT) poses are generated using a Livox Mid$360$ LiDAR with the Fast-LIO$2$ algorithm \cite{xu2022fast}.
All learning-based models were used without fine-tuning.

\begin{table}[t]
  \centering
  \begin{threeparttable}
    \caption{Absolute trajectory error on the simulated sequences.}
    \vspace{-0.2cm}
    \renewcommand\arraystretch{1.0}
    \renewcommand\tabcolsep{1.8pt}
    \scriptsize
    \begin{tabular}{c|c|c|cccc}
      \toprule[0.03cm]
      \textbf{Env.}                    & \textbf{Map Size}\tnote{$^*$}              & \textbf{Seq. (Length)} & \textbf{ICP ($15$)} & \textbf{VLoc ($1$)} & \textbf{PS ($15$)} & \textbf{PS-Opt ($1$)} \\
      \midrule[0.03cm]
      \multirow{3}{*}{\textit{Env}$0$} & \multirow{3}{*}{\begin{tabular}[c]{@{}c@{}}$|\mathcal{N}|=20$ \\$|\mathcal{E}|=71$\\ $21$\textit{KB}$+$$1.4$\textit{MB}\end{tabular}} & Seq$0$ ($34.70m$)      & $1.400$             & $\mathbf{0.013}$    & $0.045$            & $0.036$               \\
                                       &                                            & Seq$1$ ($27.10m$)      & $0.836$             & $\mathbf{0.016}$    & $0.049$            & $0.090$               \\
                                       &                                            & Seq$2$ ($24.40m$)      & $0.747$             & $\mathbf{0.010}$    & $0.080$            & $0.051$               \\
      \midrule

      \multirow{3}{*}{\textit{Env}$1$} & \multirow{3}{*}{\begin{tabular}[c]{@{}c@{}}$|\mathcal{N}|=26$\\ $|\mathcal{E}|=104$\\ $27$\textit{KB}$+$$2.3$\textit{MB}\end{tabular}} & Seq$3$ ($51.10m$)      & $2.031$             & $\mathbf{0.029}$    & $0.065$            & $0.057$               \\
                                       &                                            & Seq$4$ ($38.70m$)      & $0.173$             & $\mathbf{0.042}$    & $0.043$            & $0.039$               \\

                                       &                                            & Seq$5$ ($58.30m$)      & $4.589$             & $\mathbf{0.025}$    & $0.104$            & $0.075$               \\
      \midrule
      \multirow{3}{*}{\textit{Env}$2$} & \multirow{3}{*}{\begin{tabular}[c]{@{}c@{}}$|\mathcal{N}|=63$ \\ $|\mathcal{E}|=242$\\ $65$\textit{KB}$+$$4.5$\textit{MB}\end{tabular}} & Seq$6$ ($206.5m$)      & $22.921$            & $\mathbf{0.042}$    & $0.143$            & $0.134$               \\
                                       &                                            & Seq$7$ ($124.9m$)      & $5.716$             & $\mathbf{0.055}$    & $0.757$            & $0.295$               \\
                                       &                                            & Seq$8$ ($182.7m$)      & $6.953$             & $\mathbf{0.073}$    & $2.480$            & $0.926$               \\
      \bottomrule[0.03cm]
    \end{tabular}
    \begin{tablenotes}
      \footnotesize
      \item[*]Number of nodes, edges, and storage of VPR descriptors and images.
    \end{tablenotes}
    \label{tab:experiment_vloc_simu}
  \end{threeparttable}
\end{table}

\subsection{Map-free Relocalization}
\label{sec:experiment_rpe}
\subsubsection{Dataset}
Related datasets, such as the map-free relocalization benchmark \cite{arnold2022map} and several SfM datasets, have limitations compared to ideal robot datasets. They typically focus on a specific object in circular motion, whereas robots often navigate in straight paths and operate in unstructured environments.
To address this, we propose a new dataset for this task, including both simulated and real-world data:
\textit{(1)} \textbf{Simulated Matterport3D} \cite{chang2017matterport3d}: An indoor dataset featuring similar scenes and occluded viewpoint, with $27$ reference images and $703$ query images.
\textit{(2)} \textbf{UCL Campus}: Captured by the ANYmal-D quadruped robot between two campus buildings. The environment is crowded and structureless, with $9$ reference images and $290$ query images.
\textit{(3)} \textbf{HKUSTGZ Campus}: Generaetd from sequences presented in \cite{he2024accurate}, with depth provided by LiDAR map processing. This dataset features significant appearance, viewpoint, and position changes, with $30$ reference images and $1897$ query images.
Several examples are shown in Fig. \ref{fig:exp_rpe_benchmark}.

\subsubsection{Performance}
We report the results of relative pose estimation in Table \ref{tab:pose_error_evaluation}.
These metrics reflect different aspects of a method's performance:
maximum error and success rate indicate reliability,
median error reflects accuracy,
precision of pose error within $[Xcm,Y^{\circ}]$ shows the error bound and robustness, and time cost reflects efficiency.
MASt3R \cite{leroy2024grounding} outperforms other matchers across all datasets while maintaining competitive computation time compared to its closest competitors, DUSt3R \cite{leroy2024grounding} and GIM-DKM \cite{shen2024gim,edstedt2023dkm}. Therefore, we select MASt3R as the matcher for VLoc.

\begin{table}[t]
  \centering
  \begin{threeparttable}
    \caption{Absolute trajectory error on the real-world sequences.}
    \vspace{-0.2cm}
    \renewcommand\arraystretch{1.0}
    \renewcommand\tabcolsep{4.3pt}
    \scriptsize
    \begin{tabular}{c|ccc}
      \toprule[0.03cm]
      \textbf{Map Size}        & \multicolumn{3}{c}{$|\mathcal{N}|=81, |\mathcal{E}|=383$, Storage: $83$\textit{KB}$+$$17$\textit{MB}}                                                                                           \\
      \midrule
      \textbf{Methods}         & \textbf{Seq$\mathbf{0}$ ($\mathbf{198m}$)}                                                            & \textbf{Seq$\mathbf{1}$ ($\mathbf{172m}$)} & \textbf{Seq$\mathbf{2}$ ($\mathbf{129m}$)} \\
      \midrule[0.03cm]
      RobotOdom ($20$)         & $1.193$                                                                                               & $1.129$                                    & $0.536$                                    \\
      RobotOdom-VLoc ($0.5$)   & $\mathbf{0.238}$                                                                                      & $\mathbf{0.217}$                           & $0.191$                                    \\
      RobotOdom-PS ($20$)      & $0.548$                                                                                               & $0.557$                                    & $0.300$                                    \\
      RobotOdom-PS-Opt ($0.5$) & $\mathbf{0.238}$                                                                                      & $0.242$                                    & $\mathbf{0.170}$                           \\
      \midrule
      AirSLAM ($10$)           & $43.483$                                                                                              & $20.881$                                   & $17.704$                                   \\
      AirSLAM-VLoc ($0.5$)     & $\mathbf{0.233}$                                                                                      & $\mathbf{0.226}$                           & $\mathbf{0.152}$                           \\
      AirSLAM-PS ($10$)        & $8.791$                                                                                               & $5.231$                                    & $1.040$                                    \\
      AirSLAM-PS-Opt ($0.5$)   & $2.913$                                                                                               & $1.101$                                    & $0.386$                                    \\
      \bottomrule[0.03cm]
    \end{tabular}
    \label{tab:experiment_vloc_real}
    \vspace{-0.5cm}
  \end{threeparttable}
\end{table}



\subsection{Visual Localization}
\label{sec:experiment_vloc}
\subsubsection{Dataset}
We generate $3$ simulated sequences for each of $3$ Matterport3D environments using the autonomous navigation stack \cite{cao2022autonomous} (given a set of waypoints): \textit{17DRP5sb8fy}, \textit{EDJbREhghzL}, and \textit{B6ByNegPMK} (denoted as \textit{Simu-Seq}$0$-$8$ for \textit{Env}$0$-$2$), with increasing environment size. Some sequences include motion into regions not covered by the map can be used to show the feasibility of our method in unseen areas.
We also manually control the robot to collect $3$ real-world outdoor sequences, where some regions are crowded with people and others on a bridge are structureless. \textit{Simu-Seq}$0$, \textit{Simu-Seq}$3$, \textit{Simu-Seq}$6$, and \textit{Real-Seq}$0$, containing geo-tagged images, are used to build the topo-metric map.

\subsubsection{Performance}
\begin{figure}[t]
  \begin{center}
    \includegraphics[width=0.97\linewidth]{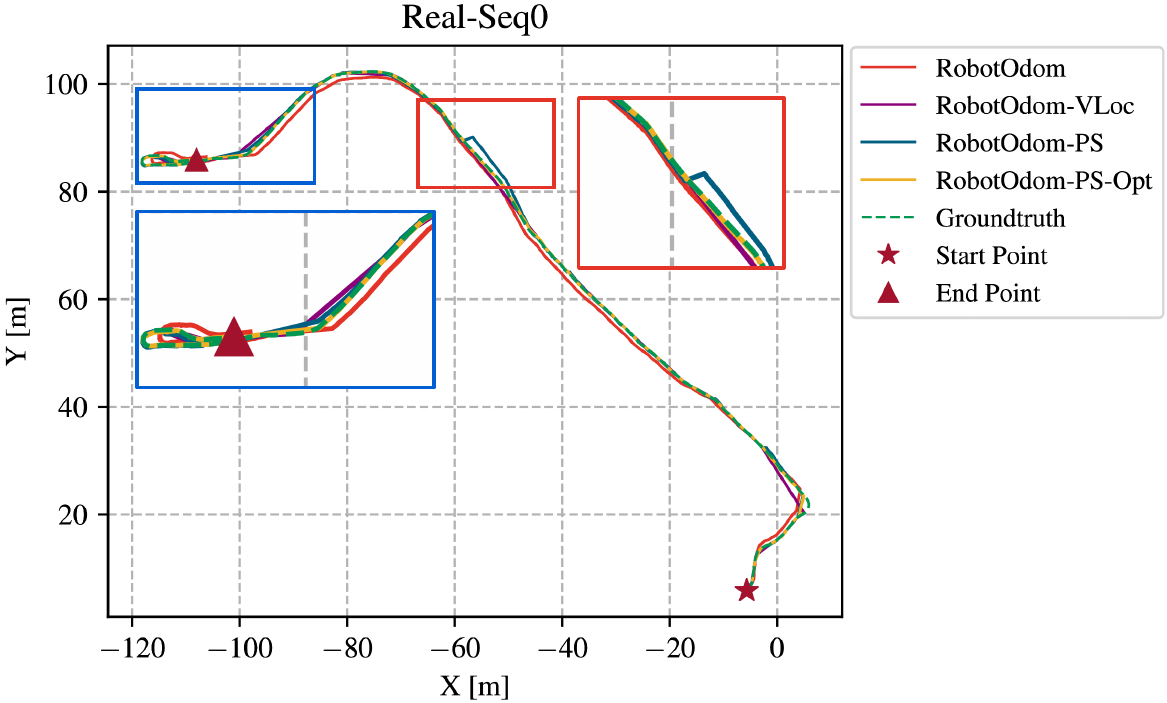}
  \end{center}
  \vspace{-0.35cm}
  \caption{Estimated trajectories on a real-world sequence traversing a bridge. VLoc's trajectory closely aligns with the ground truth, demonstrating high accuracy. The sudden changes in both \textit{RobotOdom} and \textit{RobotOdom-PS} are due to corrections from VLoc's results.}
  \label{fig:exp_real_seq0_traj}
  \vspace{-0.5cm}
\end{figure}

We report the absolute trajectory error (ATE) of modules in LiteVLoc in Tables \ref{tab:experiment_vloc_simu} and \ref{tab:experiment_vloc_real}, along with a description of the map.
The number following each method indicates the frequency of its output.
In Table \ref{tab:experiment_vloc_real}, \textit{RobotOdom} refers to the proprioceptive odometry provided by the robot, and \textit{X-PS-Opt} denotes the batch optimization results, where historical poses are optimized using additional measurements. We also replace \textit{RobotOdom} with AirSLAM stereo camera odometry \cite{xu2024airsladm} for similar experiments.
In Fig. \ref{fig:exp_real_seq0_traj}, we visualize the trajectories of our method on \textit{Real-Seq0}.
These results confirm our hypothesis:
\textit{(1)} low-rate VLoc can effectively correct the accumulative drift from local odometry,
\textit{(2)} the localization accuracy is sufficient for navigation, and
\textit{(3)} batch optimization results provide good initial guess for accurate mapping missions.





\begin{figure*}[t]
  \begin{center}
    \includegraphics[width=0.91\linewidth]{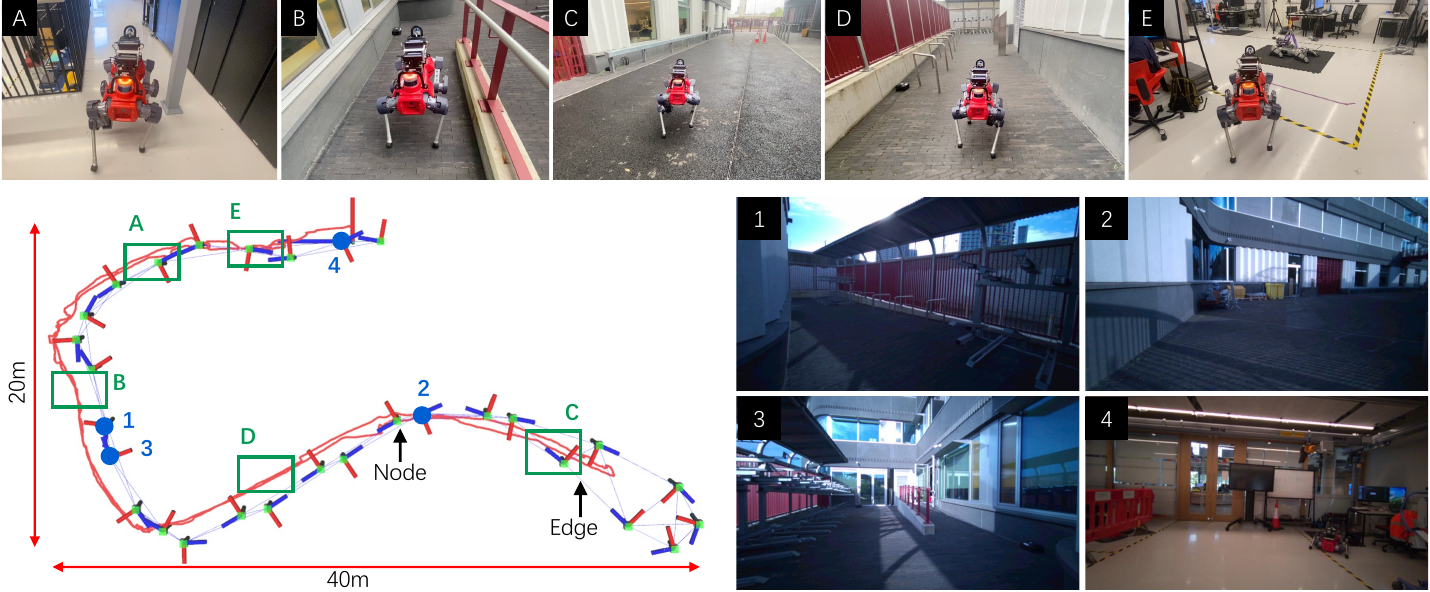}
  \end{center}
  \vspace{-0.4cm}
  \caption{Real-world experiment with a legged robot, with the red curve showning the robot's trajectory estimated by our VLoc method. The nodes (visualized as axes: \rt{red}-X, \gt{green}-Y, and \bt{blue}-Z) and connected edges indicate the topo-metric map's structure. The robot is guided by a sequence of goal images that were captured by the AR glasses (right). It starts inside a room, navigates outside, follows a circular route, and returns.
    It totally traverses $94.5m$ in $314s$ with an average speed of $0.3m/s$.
    Green boxes A-E highlight key planning events along the route: (A) a narrow pathway, (B) descending slope, (C and D) outdoor pathways, and (E) the lab region. Refer to the video for a full demonstration of the navigation process.}
  \label{fig:real_world_vnav}
  \vspace{-0.4cm}
\end{figure*}

\subsection{Image Goal Visual Navigation}
\label{sec:experiment_vnav}
We evaluate the VNav system using three simulated tests and one real-world test. In each experiment, the robot is sequentially given five goal images and tasked with completing the navigation process. Table \ref{tab:experiment_vnav_simu} presents the results of the simulated experiments, comparing the VNav system to the SoTA LiDAR-based navigation system \cite{cao2022autonomous} (GT localization $+$ Falco) as a baseline.
Due to the limited field of view of the camera, VNav sometimes encounters local minima, requiring the robot to rotate more frequently.
This increases the time taken to reach the goal.
However, the length of the navigation path does not increase significantly, indicating that VLoc consistently tracks the robot's position accurately.

Fig. \ref{fig:real_world_vnav} shows the results of the VNav system in real-world environments.
We use the AR glasses \cite{engel2023project} that provide geo-tagged images for the map construction.
The real-world test further demonstrates the flexibility of our method, even in the presence of domain gaps and viewpoint differences between the reference images and the robot's observations (see Fig. \ref{fig:exp_rpe_benchmark} and \ref{fig:real_world_vnav}).
The robot successfully navigates from an lab to an outdoor pathway, guided by a sequence of goal images.
It is also capable of autonomously navigating through narrow doorways and avoiding obstacles.
The $94.5m$-route was completed in $314s$, with an average speed of $0.3m/s$.
The indoor and outdoor environments present different lighting conditions, resulting in varying noise levels in the RGB and depth measurements.
A video of the experiment provides a clearer visualization of this process.

\begin{table}[t]
  \centering
  \caption{Navigation performance in simulated environments.}
  \renewcommand\arraystretch{1.0}
  \renewcommand\tabcolsep{3.7pt}
  \scriptsize
  \begin{tabular}{c|ccc}
    \toprule[0.03cm]
                                      & \multicolumn{3}{c}{Navigation Time $[s]$ and Path Length $[m]$}                                                                             \\
    \midrule
    \multirow{1}{*}{\textbf{Methods}} & \multicolumn{1}{c}{\textit{Env}$0$}                             & \multicolumn{1}{c}{\textit{Env}$1$} & \multicolumn{1}{c}{\textit{Env}$2$} \\

    \midrule[0.03cm]
    GTLoc $+$ Falco \cite{zhang2020falco}                     & $40.3s/25.5m$                                                   & $42.7s/28.3m$                       & $266.9s/220.7m$                     \\
    LiteVLoc $+$ iPlanner \cite{yang2023iplanner}                  & $50.0s/28.6m$                                                   & $52.9s/29.1m$                       & $280.3s/223.7m$                     \\
    \bottomrule[0.03cm]
  \end{tabular}
  \label{tab:experiment_vnav_simu}
  \vspace{-0.3cm}
\end{table}

\section{Conclusion}
\label{sec:conclusion}
This paper proposes a simple yet effective VLoc method using topo-metric maps for robot navigation.
The map-lite feature of ours is different from mainstream methods that using detailed 3D maps to represent environments.
Experiments demonstrate that while the proposed LiteVLoc has much room for improvement, it can be used for many non-safety-critical applications.
In addition, LiteVLoc reduces the strict mapping requirements for navigation.
As spatially aware mobile devices become more prevalent, the gap between scalable navigation and real-world deployment will be narrowed.
Future work will focus on optimizing real-time performance and further integrating with downstream navigation to handle more complex tasks in dynamic environments for efficient path planning~\cite{kanoulas2019curved, liu2023legged, liu2024dipper, stamatopoulou2024dippest}, using advanced SLAM methods~\cite{kanoulas2018rxkinfu, jiao2024RTMapping, cheng2024LoGS}.


\bibliographystyle{IEEEtran}
\bibliography{bible}

\begin{thebibliography}{10}
\providecommand{\url}[1]{#1}
\csname url@samestyle\endcsname
\providecommand{\newblock}{\relax}
\providecommand{\bibinfo}[2]{#2}
\providecommand{\BIBentrySTDinterwordspacing}{\spaceskip=0pt\relax}
\providecommand{\BIBentryALTinterwordstretchfactor}{4}
\providecommand{\BIBentryALTinterwordspacing}{\spaceskip=\fontdimen2\font plus
\BIBentryALTinterwordstretchfactor\fontdimen3\font minus \fontdimen4\font\relax}
\providecommand{\BIBforeignlanguage}[2]{{%
\expandafter\ifx\csname l@#1\endcsname\relax
\typeout{** WARNING: IEEEtran.bst: No hyphenation pattern has been}%
\typeout{** loaded for the language `#1'. Using the pattern for}%
\typeout{** the default language instead.}%
\else
\language=\csname l@#1\endcsname
\fi
#2}}
\providecommand{\BIBdecl}{\relax}
\BIBdecl

\bibitem{noh2017large}
H.~Noh, A.~Araujo, J.~Sim, T.~Weyand, and B.~Han, ``Large-scale image retrieval with attentive deep local features,'' in \emph{Proceedings of the IEEE international conference on computer vision}, 2017, pp. 3456--3465.

\bibitem{dusmanu2019d2}
M.~Dusmanu, I.~Rocco, T.~Pajdla, M.~Pollefeys, J.~Sivic, A.~Torii, and T.~Sattler, ``{D2-Net}: A trainable cnn for joint description and detection of local features,'' in \emph{Proceedings of the ieee/cvf conference on computer vision and pattern recognition}, 2019, pp. 8092--8101.

\bibitem{sattler2016efficient}
T.~Sattler, B.~Leibe, and L.~Kobbelt, ``Efficient and effective prioritized matching for large-scale image-based localization,'' \emph{IEEE transactions on pattern analysis and machine intelligence}, vol.~39, no.~9, pp. 1744--1756, 2016.

\bibitem{taira2018inloc}
H.~Taira, M.~Okutomi, T.~Sattler, M.~Cimpoi, M.~Pollefeys, J.~Sivic, T.~Pajdla, and A.~Torii, ``Inloc: Indoor visual localization with dense matching and view synthesis,'' in \emph{Proceedings of the IEEE Conference on Computer Vision and Pattern Recognition}, 2018, pp. 7199--7209.

\bibitem{sarlin2019coarse}
P.-E. Sarlin, C.~Cadena, R.~Siegwart, and M.~Dymczyk, ``From coarse to fine: Robust hierarchical localization at large scale,'' in \emph{CVPR}, 2019.

\bibitem{schoenberger2016sfm}
J.~L. Sch\"{o}nberger and J.-M. Frahm, ``Structure-from-motion revisited,'' in \emph{Conference on Computer Vision and Pattern Recognition (CVPR)}, 2016.

\bibitem{brachmann2023accelerated}
\BIBentryALTinterwordspacing
E.~Brachmann, T.~Cavallari, and V.~A. Prisacariu, ``Accelerated coordinate encoding: Learning to relocalize in minutes using rgb and poses,'' \emph{2023 IEEE/CVF Conference on Computer Vision and Pattern Recognition (CVPR)}, pp. 5044--5053, 2023. [Online]. Available: \url{https://api.semanticscholar.org/CorpusID:258841110}
\BIBentrySTDinterwordspacing

\bibitem{yin2024general}
P.~Yin, J.~Jiao, S.~Zhao, L.~Xu, G.~Huang, H.~Choset, S.~Scherer, and J.~Han, ``General place recognition survey: Towards real-world autonomy,'' \emph{arXiv preprint arXiv:2405.04812}, 2024.

\bibitem{dayoub2013vision}
F.~Dayoub, T.~Morris, B.~Upcroft, and P.~Corke, ``Vision-only autonomous navigation using topometric maps,'' in \emph{2013 IEEE/RSJ International Conference on Intelligent Robots and Systems}.\hskip 1em plus 0.5em minus 0.4em\relax IEEE, 2013, pp. 1923--1929.

\bibitem{roussel2020deep}
T.~Roussel, P.~Chakravarty, G.~Pandey, T.~Tuytelaars, and L.~Van~Eycken, ``Deep-geometric 6 dof localization from a single image in topo-metric maps,'' \emph{arXiv preprint arXiv:2002.01210}, 2020.

\bibitem{rosinol2021kimera}
A.~Rosinol, A.~Violette, M.~Abate, N.~Hughes, Y.~Chang, J.~Shi, A.~Gupta, and L.~Carlone, ``Kimera: From slam to spatial perception with 3d dynamic scene graphs,'' \emph{The International Journal of Robotics Research}, vol.~40, no. 12-14, pp. 1510--1546, 2021.

\bibitem{he2023metric}
Y.~He, I.~Fang, Y.~Li, R.~B. Shah, and C.~Feng, ``Metric-free exploration for topological mapping by task and motion imitation in feature space,'' in \emph{Proceedings of Robotics: Science and Systems}, Daegu, Republic of Korea, July 2023.

\bibitem{kim2023topological}
N.~Kim, O.~Kwon, H.~Yoo, Y.~Choi, J.~Park, and S.~Oh, ``Topological semantic graph memory for image-goal navigation,'' in \emph{Conference on Robot Learning}.\hskip 1em plus 0.5em minus 0.4em\relax PMLR, 2023, pp. 393--402.

\bibitem{hughes2024foundations}
N.~Hughes, Y.~Chang, S.~Hu, R.~Talak, R.~Abdulhai, J.~Strader, and L.~Carlone, ``Foundations of spatial perception for robotics: Hierarchical representations and real-time systems,'' \emph{The International Journal of Robotics Research}, p. 02783649241229725, 2024.

\bibitem{arnold2022map}
E.~Arnold, J.~Wynn, S.~Vicente, G.~Garcia-Hernando, A.~Monszpart, V.~Prisacariu, D.~Turmukhambetov, and E.~Brachmann, ``Map-free visual relocalization: Metric pose relative to a single image,'' in \emph{European Conference on Computer Vision}.\hskip 1em plus 0.5em minus 0.4em\relax Springer, 2022, pp. 690--708.

\bibitem{engel2023project}
J.~Engel, K.~Somasundaram, M.~Goesele, A.~Sun, A.~Gamino, A.~Turner, A.~Talattof, A.~Yuan, B.~Souti, B.~Meredith \emph{et~al.}, ``Project aria: A new tool for egocentric multi-modal ai research,'' \emph{arXiv preprint arXiv:2308.13561}, 2023.

\bibitem{arandjelovi2015netvlad}
\BIBentryALTinterwordspacing
R.~Arandjelovi{\'c}, P.~Gron{\'a}t, A.~Torii, T.~Pajdla, and J.~Sivic, ``Netvlad: Cnn architecture for weakly supervised place recognition,'' \emph{2016 IEEE Conference on Computer Vision and Pattern Recognition (CVPR)}, pp. 5297--5307, 2015. [Online]. Available: \url{https://api.semanticscholar.org/CorpusID:44604205}
\BIBentrySTDinterwordspacing

\bibitem{xu2021probabilisticvp}
\BIBentryALTinterwordspacing
M.~Xu, N.~S{\"u}nderhauf, and M.~Milford, ``Probabilistic visual place recognition for hierarchical localization,'' \emph{IEEE Robotics Autom. Lett.}, vol.~6, pp. 311--318, 2021. [Online]. Available: \url{https://api.semanticscholar.org/CorpusID:229643627}
\BIBentrySTDinterwordspacing

\bibitem{berton2022rethinking}
G.~Berton, C.~Masone, and B.~Caputo, ``Rethinking visual geo-localization for large-scale applications,'' in \emph{Proceedings of the IEEE/CVF Conference on Computer Vision and Pattern Recognition}, 2022, pp. 4878--4888.

\bibitem{berton2023eigenplaces}
G.~Berton, G.~Trivigno, B.~Caputo, and C.~Masone, ``Eigenplaces: Training viewpoint robust models for visual place recognition,'' in \emph{Proceedings of the IEEE/CVF International Conference on Computer Vision}, 2023, pp. 11\,080--11\,090.

\bibitem{gordo2017end}
A.~Gordo, J.~Almazan, J.~Revaud, and D.~Larlus, ``End-to-end learning of deep visual representations for image retrieval,'' \emph{International Journal of Computer Vision}, vol. 124, no.~2, pp. 237--254, 2017.

\bibitem{sattler2019understanding}
\BIBentryALTinterwordspacing
T.~Sattler, Q.~Zhou, M.~Pollefeys, and L.~Leal-Taix{\'e}, ``Understanding the limitations of cnn-based absolute camera pose regression,'' \emph{2019 IEEE/CVF Conference on Computer Vision and Pattern Recognition (CVPR)}, pp. 3297--3307, 2019. [Online]. Available: \url{https://api.semanticscholar.org/CorpusID:81979654}
\BIBentrySTDinterwordspacing

\bibitem{dong2023lazy}
S.~Dong, S.~Liu, H.~Guo, B.~Chen, and M.~Pollefeys, ``Lazy visual localization via motion averaging,'' \emph{arXiv preprint arXiv:2307.09981}, 2023.

\bibitem{zaffar2024estimation}
M.~Zaffar, L.~Nan, and J.~F. Kooij, ``On the estimation of image-matching uncertainty in visual place recognition,'' in \emph{Proceedings of the IEEE/CVF Conference on Computer Vision and Pattern Recognition}, 2024, pp. 17\,743--17\,753.

\bibitem{kendall2015posenet}
A.~Kendall, M.~Grimes, and R.~Cipolla, ``Posenet: A convolutional network for real-time 6-dof camera relocalization,'' in \emph{Proceedings of the IEEE international conference on computer vision}, 2015, pp. 2938--2946.

\bibitem{kendall2017geometric}
A.~Kendall and R.~Cipolla, ``Geometric loss functions for camera pose regression with deep learning,'' in \emph{IEEE conference on computer vision and pattern recognition}, 2017, pp. 5974--5983.

\bibitem{shavit2021learning}
Y.~Shavit, R.~Ferens, and Y.~Keller, ``Learning multi-scene absolute pose regression with transformers,'' in \emph{IEEE/CVF International Conference on Computer Vision}, 2021, pp. 2733--2742.

\bibitem{chen2022dfnet}
S.~Chen, X.~Li, Z.~Wang, and V.~A. Prisacariu, ``Dfnet: Enhance absolute pose regression with direct feature matching,'' in \emph{ECCV 2022. Tel Aviv, Israel, October 23--27, 2022, Part X}.\hskip 1em plus 0.5em minus 0.4em\relax Springer, 2022.

\bibitem{balntas2018relocnet}
V.~Balntas, S.~Li, and V.~Prisacariu, ``Relocnet: Continuous metric learning relocalisation using neural nets,'' in \emph{Proceedings of the European Conference on Computer Vision (ECCV)}, 2018, pp. 751--767.

\bibitem{laskar2017camera}
Z.~Laskar, I.~Melekhov, S.~Kalia, and J.~Kannala, ``Camera relocalization by computing pairwise relative poses using convolutional neural network,'' in \emph{Proceedings of the IEEE International Conference on Computer Vision Workshops}, 2017, pp. 929--938.

\bibitem{10610903}
C.~Liu, S.~Chen, Y.~Zhao, H.~Huang, V.~Prisacariu, and T.~Braud, ``Hr-apr: Apr-agnostic framework with uncertainty estimation and hierarchical refinement for camera relocalisation,'' in \emph{2024 IEEE International Conference on Robotics and Automation (ICRA)}, 2024, pp. 8544--8550.

\bibitem{detone2018superpoint}
D.~DeTone, T.~Malisiewicz, and A.~Rabinovich, ``Superpoint: Self-supervised interest point detection and description,'' in \emph{Proceedings of the IEEE conference on computer vision and pattern recognition workshops}, 2018, pp. 224--236.

\bibitem{lindenberger2023lightglue}
P.~Lindenberger, P.-E. Sarlin, and M.~Pollefeys, ``{LightGlue}: Local feature matching at light speed,'' in \emph{Proceedings of the IEEE/CVF International Conference on Computer Vision}, 2023, pp. 17\,627--17\,638.

\bibitem{sarlin2020superglue}
P.-E. Sarlin, D.~DeTone, T.~Malisiewicz, and A.~Rabinovich, ``Superglue: Learning feature matching with graph neural networks,'' in \emph{Proceedings of the IEEE/CVF conference on computer vision and pattern recognition}, 2020, pp. 4938--4947.

\bibitem{sun2021loftr}
J.~Sun, Z.~Shen, Y.~Wang, H.~Bao, and X.~Zhou, ``{LoFTR}: Detector-free local feature matching with transformers,'' in \emph{Proceedings of the IEEE/CVF conference on computer vision and pattern recognition}, 2021, pp. 8922--8931.

\bibitem{gao2003complete}
\BIBentryALTinterwordspacing
X.~Gao, X.~Hou, J.~Tang, and H.-F. Cheng, ``Complete solution classification for the perspective-three-point problem,'' \emph{IEEE Trans. Pattern Anal. Mach. Intell.}, vol.~25, pp. 930--943, 2003. [Online]. Available: \url{https://api.semanticscholar.org/CorpusID:15869446}
\BIBentrySTDinterwordspacing

\bibitem{panek2022meshloc}
\BIBentryALTinterwordspacing
V.~Panek, Z.~Kukelova, and T.~Sattler, ``Meshloc: Mesh-based visual localization,'' in \emph{European Conference on Computer Vision}, 2022. [Online]. Available: \url{https://api.semanticscholar.org/CorpusID:251018193}
\BIBentrySTDinterwordspacing

\bibitem{moreau2023crossfirecr}
\BIBentryALTinterwordspacing
A.~Moreau, N.~Piasco, M.~Bennehar, D.~V. Tsishkou, B.~Stanciulescu, and A.~de~La~Fortelle, ``Crossfire: Camera relocalization on self-supervised features from an implicit representation,'' \emph{2023 IEEE/CVF International Conference on Computer Vision (ICCV)}, pp. 252--262, 2023. [Online]. Available: \url{https://api.semanticscholar.org/CorpusID:257427144}
\BIBentrySTDinterwordspacing

\bibitem{liu2023nerfloc}
\BIBentryALTinterwordspacing
J.~Liu, Q.~Nie, Y.~Liu, and C.~Wang, ``Nerf-loc: Visual localization with conditional neural radiance field,'' \emph{2023 IEEE International Conference on Robotics and Automation (ICRA)}, pp. 9385--9392, 2023. [Online]. Available: \url{https://api.semanticscholar.org/CorpusID:258180042}
\BIBentrySTDinterwordspacing

\bibitem{zhou2024nerfect}
Q.~Zhou, M.~Maximov, O.~Litany, and L.~Leal-Taix{\'e}, ``The nerfect match: Exploring nerf features for visual localization,'' \emph{arXiv preprint arXiv:2403.09577}, 2024.

\bibitem{liu2024gsloc}
\BIBentryALTinterwordspacing
C.~Liu, S.~Chen, Y.~Bhalgat, S.~Hu, Z.~Wang, M.~Cheng, V.~A. Prisacariu, and T.~Braud, ``Gsloc: Efficient camera pose refinement via 3d gaussian splatting,'' 2024. [Online]. Available: \url{https://api.semanticscholar.org/CorpusID:271916288}
\BIBentrySTDinterwordspacing

\bibitem{brachmann2016dsac}
\BIBentryALTinterwordspacing
E.~Brachmann, A.~Krull, S.~Nowozin, J.~Shotton, F.~Michel, S.~Gumhold, and C.~Rother, ``Dsac — differentiable ransac for camera localization,'' \emph{2017 IEEE Conference on Computer Vision and Pattern Recognition (CVPR)}, pp. 2492--2500, 2016. [Online]. Available: \url{https://api.semanticscholar.org/CorpusID:4001530}
\BIBentrySTDinterwordspacing

\bibitem{brachmann2021dsacstar}
E.~Brachmann and C.~Rother, ``Visual camera re-localization from {RGB} and {RGB-D} images using {DSAC},'' \emph{TPAMI}, 2021.

\bibitem{brachmann2018lessmore}
------, ``Learning less is more - {6D} camera localization via {3D} surface regression,'' in \emph{CVPR}, 2018.

\bibitem{sheffer2020pnp}
R.~Sheffer and A.~Wiesel, ``Pnp-net: A hybrid perspective-n-point network,'' \emph{arXiv preprint arXiv:2003.04626}, 2020.

\bibitem{barfoot2024state}
T.~D. Barfoot, \emph{State estimation for robotics}.\hskip 1em plus 0.5em minus 0.4em\relax Cambridge University Press, 2024.

\bibitem{xu2024d}
H.~Xu, P.~Liu, X.~Chen, and S.~Shen, ``{D2SLAM}: Decentralized and distributed collaborative visual-inertial slam system for aerial swarm,'' \emph{IEEE Transactions on Robotics}, 2024.

\bibitem{liu2024marvin}
C.~Liu, Y.~Zhao, and T.~Braud, ``Marvin: Mobile ar dataset with visual-inertial data,'' in \emph{2024 IEEE Conference on Virtual Reality and 3D User Interfaces Abstracts and Workshops (VRW)}.\hskip 1em plus 0.5em minus 0.4em\relax IEEE, 2024, pp. 532--538.

\bibitem{shah2023gnm}
D.~Shah, A.~Sridhar, A.~Bhorkar, N.~Hirose, and S.~Levine, ``{GNM}: A general navigation model to drive any robot,'' in \emph{2023 IEEE International Conference on Robotics and Automation (ICRA)}.\hskip 1em plus 0.5em minus 0.4em\relax IEEE, 2023, pp. 7226--7233.

\bibitem{anguelov2010google}
D.~Anguelov, C.~Dulong, D.~Filip, C.~Frueh, S.~Lafon, R.~Lyon, A.~Ogale, L.~Vincent, and J.~Weaver, ``Google street view: Capturing the world at street level,'' \emph{Computer}, vol.~43, no.~6, pp. 32--38, 2010.

\bibitem{jiao2021greedy}
J.~Jiao, Y.~Zhu, H.~Ye, H.~Huang, P.~Yun, L.~Jiang, L.~Wang, and M.~Liu, ``Greedy-based feature selection for efficient lidar slam,'' in \emph{2021 IEEE International Conference on Robotics and Automation (ICRA)}.\hskip 1em plus 0.5em minus 0.4em\relax IEEE, 2021, pp. 5222--5228.

\bibitem{he2016deep}
K.~He, X.~Zhang, S.~Ren, and J.~Sun, ``Deep residual learning for image recognition,'' in \emph{Proceedings of the IEEE conference on computer vision and pattern recognition}, 2016, pp. 770--778.

\bibitem{leroy2024grounding}
V.~Leroy, Y.~Cabon, and J.~Revaud, ``Grounding image matching in 3d with {MASt3R},'' \emph{arXiv preprint arXiv:2406.09756}, 2024.

\bibitem{fischler1981random}
M.~A. Fischler and R.~C. Bolles, ``Random sample consensus: a paradigm for model fitting with applications to image analysis and automated cartography,'' \emph{Communications of the ACM}, vol.~24, no.~6, pp. 381--395, 1981.

\bibitem{rusinkiewicz2001efficient}
S.~Rusinkiewicz and M.~Levoy, ``Efficient variants of the icp algorithm,'' in \emph{Proceedings third international conference on 3-D digital imaging and modeling}.\hskip 1em plus 0.5em minus 0.4em\relax IEEE, 2001, pp. 145--152.

\bibitem{bloesch2017state}
M.~Bloesch, ``State estimation for legged robots-kinematics, inertial sensing, and computer vision,'' Ph.D. dissertation, ETH Zurich, 2017.

\bibitem{framk2022gtsam}
\BIBentryALTinterwordspacing
F.~Dellaert and G.~Contributors, ``borglab/gtsam,'' May 2022. [Online]. Available: \url{https://github.com/borglab/gtsam)}
\BIBentrySTDinterwordspacing

\bibitem{yang2023iplanner}
F.~Yang, C.~Wang, C.~Cadena, and M.~Hutter, ``iplanner: Imperative path planning,'' \emph{arXiv preprint arXiv:2302.11434}, 2023.

\bibitem{zhang2020falco}
J.~Zhang, C.~Hu, R.~G. Chadha, and S.~Singh, ``Falco: Fast likelihood-based collision avoidance with extension to human-guided navigation,'' \emph{Journal of Field Robotics}, vol.~37, no.~8, pp. 1300--1313, 2020.

\bibitem{edstedt2024roma}
J.~Edstedt, Q.~Sun, G.~B{\"o}kman, M.~Wadenb{\"a}ck, and M.~Felsberg, ``{RoMa}: Robust dense feature matching,'' in \emph{Proceedings of the IEEE/CVF Conference on Computer Vision and Pattern Recognition}, 2024, pp. 19\,790--19\,800.

\bibitem{wang2024dust3r}
S.~Wang, V.~Leroy, Y.~Cabon, B.~Chidlovskii, and J.~Revaud, ``{Dust3r}: Geometric 3d vision made easy,'' in \emph{Proceedings of the IEEE/CVF Conference on Computer Vision and Pattern Recognition}, 2024, pp. 20\,697--20\,709.

\bibitem{wang2024efficient}
Y.~Wang, X.~He, S.~Peng, D.~Tan, and X.~Zhou, ``{Efficient LoFTR}: Semi-dense local feature matching with sparse-like speed,'' in \emph{Proceedings of the IEEE/CVF Conference on Computer Vision and Pattern Recognition}, 2024, pp. 21\,666--21\,675.

\bibitem{wang2022matchformer}
Q.~Wang, J.~Zhang, K.~Yang, K.~Peng, and R.~Stiefelhagen, ``{MatchFormer}: Interleaving attention in transformers for feature matching,'' in \emph{Proceedings of the Asian Conference on Computer Vision}, 2022, pp. 2746--2762.

\bibitem{potje2024xfeat}
G.~Potje, F.~Cadar, A.~Araujo, R.~Martins, and E.~R. Nascimento, ``{XFeat}: Accelerated features for lightweight image matching,'' in \emph{Proceedings of the IEEE/CVF Conference on Computer Vision and Pattern Recognition}, 2024, pp. 2682--2691.

\bibitem{shen2024gim}
X.~Shen, Z.~Cai, W.~Yin, M.~M{\"u}ller, Z.~Li, K.~Wang, X.~Chen, and C.~Wang, ``Gim: Learning generalizable image matcher from internet videos,'' \emph{arXiv preprint arXiv:2402.11095}, 2024.

\bibitem{edstedt2023dkm}
J.~Edstedt, I.~Athanasiadis, M.~Wadenb{\"a}ck, and M.~Felsberg, ``Dkm: Dense kernelized feature matching for geometry estimation,'' in \emph{Proceedings of the IEEE/CVF Conference on Computer Vision and Pattern Recognition}, 2023, pp. 17\,765--17\,775.

\bibitem{lowe2004distinctive}
D.~G. Lowe, ``Distinctive image features from scale-invariant keypoints,'' \emph{International journal of computer vision}, vol.~60, pp. 91--110, 2004.

\bibitem{rublee2011orb}
E.~Rublee, V.~Rabaud, K.~Konolige, and G.~Bradski, ``Orb: An efficient alternative to sift or surf,'' in \emph{2011 International conference on computer vision}.\hskip 1em plus 0.5em minus 0.4em\relax Ieee, 2011, pp. 2564--2571.

\bibitem{Hutter2016ANYmalA}
\BIBentryALTinterwordspacing
M.~Hutter, C.~Gehring, D.~Jud, A.~Lauber, D.~Bellicoso, V.~Tsounis, J.~Hwangbo, K.~Bodie, P.~Fankhauser, M.~Bloesch, R.~Diethelm, S.~Bachmann, A.~Melzer, and M.~A. H{\"o}pflinger, ``Anymal - a highly mobile and dynamic quadrupedal robot,'' \emph{2016 IEEE/RSJ International Conference on Intelligent Robots and Systems (IROS)}, pp. 38--44, 2016. [Online]. Available: \url{https://api.semanticscholar.org/CorpusID:19007565}
\BIBentrySTDinterwordspacing

\bibitem{xu2022fast}
W.~Xu, Y.~Cai, D.~He, J.~Lin, and F.~Zhang, ``{FAST-LIO2}: Fast direct lidar-inertial odometry,'' \emph{IEEE Transactions on Robotics}, vol.~38, no.~4, pp. 2053--2073, 2022.

\bibitem{chang2017matterport3d}
A.~Chang, A.~Dai, T.~Funkhouser, M.~Halber, M.~Niessner, M.~Savva, S.~Song, A.~Zeng, and Y.~Zhang, ``Matterport3d: Learning from rgb-d data in indoor environments,'' \emph{arXiv preprint arXiv:1709.06158}, 2017.

\bibitem{he2024accurate}
J.~He, H.~Huang, S.~Zhang, J.~Jiao, C.~Liu, and M.~Liu, ``Accurate prior-centric monocular positioning with offline lidar fusion,'' in \emph{2024 IEEE International Conference on Robotics and Automation (ICRA)}.\hskip 1em plus 0.5em minus 0.4em\relax IEEE, 2024, pp. 11\,934--11\,940.

\bibitem{cao2022autonomous}
C.~Cao, H.~Zhu, F.~Yang, Y.~Xia, H.~Choset, J.~Oh, and J.~Zhang, ``Autonomous exploration development environment and the planning algorithms,'' in \emph{2022 International Conference on Robotics and Automation (ICRA)}.\hskip 1em plus 0.5em minus 0.4em\relax IEEE, 2022, pp. 8921--8928.

\bibitem{xu2024airsladm}
K.~Xu, Y.~Hao, S.~Yuan, C.~Wang, and L.~Xie, ``Airslam: An efficient and illumination-robust point-line visual slam system,'' \emph{arXiv preprint arXiv:2408.03520}, 2024.

\bibitem{kanoulas2019curved}
D.~Kanoulas, N.~G. Tsagarakis, and M.~Vona, ``Curved patch mapping and tracking for irregular terrain modeling: Application to bipedal robot foot placement,'' \emph{Robotics and Autonomous Systems}, 2019.

\bibitem{liu2023legged}
J.~Liu, S.~Lyu, D.~Hadjivelichkov, V.~Modugno, and D.~Kanoulas, ``Vit-a*: Legged robot path planning using vision transformer a*,'' in \emph{IEEE-RAS International Conference on Humanoids Robots (Humanoids)}, 2023.

\bibitem{liu2024dipper}
J.~Liu, M.~Stamatopoulou, and D.~Kanoulas, ``Dipper: Diffusion-based 2d path planner applied on legged robots,'' in \emph{2024 IEEE International Conference on Robotics and Automation (ICRA)}.\hskip 1em plus 0.5em minus 0.4em\relax IEEE, 2024, pp. 9264--9270.

\bibitem{stamatopoulou2024dippest}
M.~Stamatopoulou, J.~Liu, and D.~Kanoulas, ``Dippest: Diffusion-based path planner for synthesizing trajectories applied on quadruped robots,'' in \emph{IEEE/RSJ International Conference on Intelligent Robots and Systems (IROS)}, 2024.

\bibitem{kanoulas2018rxkinfu}
D.~Kanoulas, N.~G. Tsagarakis, and M.~Vona, ``rxkinfu: Moving volume kinectfusion for 3d perception and robotics,'' in \emph{18th IEEE/RAS International Conference on Humanoid Robots (Humanoids)}, 2018.

\bibitem{jiao2024RTMapping}
J.~Jiao, R.~Geng, Y.~Li, R.~Xin, B.~Yang, J.~Wu, L.~Wang, M.~Liu, R.~Fan, and D.~Kanoulas, ``Real-time metric-semantic mapping for autonomous navigation in outdoor environments,'' \emph{IEEE Transactions on Automation Science and Engineering (T-ASE)}, 2024.

\bibitem{cheng2024LoGS}
Y.~Cheng, J.~Jiao, Y.~Wang, and D.~Kanoulas, ``Logs: Visual localization via gaussian splatting with fewer training images,'' 2024.

\end{thebibliography}

\end{document}